\PassOptionsToPackage{table}{xcolor}
\documentclass[acmtog]{acmart}
\acmSubmissionID{1912}

\usepackage{booktabs} 
\usepackage{multirow}
\usepackage{acmart-taps}

\definecolor{best}{HTML}{FFB3B3}
\definecolor{second}{HTML}{FFD9A6}
\definecolor{third}{HTML}{FFF2A6}

\newcommand{\best}[1]{\cellcolor{best}#1}
\newcommand{\second}[1]{\cellcolor{second}#1}

\usepackage[ruled]{algorithm2e} 
\SetAlFnt{\small}
\SetAlCapFnt{\small}
\SetAlCapNameFnt{\small}
\SetAlCapHSkip{0pt}

\copyrightyear{2026}
\acmYear{2026}
\setcopyright{cc}
\setcctype{by}
\acmConference[SA Conference Papers '26]{SIGGRAPH Asia 2026 Conference Papers}{December 01--04, 2026}{Kuala Lumpur, Malaysia}
\acmBooktitle{SIGGRAPH Asia 2026 Conference Papers (SA Conference Papers '26), December 01--04, 2026, Kuala Lumpur, Malaysia}
\acmDOI{10.1145/3829340.3842283}
\acmISBN{979-8-4007-2842-6/2026/12}

\begin{document}
\title{CompMVR: Compression-Aware Multi-View Restoration Using Diffusion Models for Geometrically Consistent 3D Reconstruction}

\author{Dong-hwi Kim}
\authornote{Equal contribution}
\email{dhwi@knu.ac.kr}
\affiliation{%
  \institution{Kyungpook National University}
  \country{Republic of Korea}
}
\orcid{0000-0002-5188-8834}

\author{Chaewon Moon}
\email{moonchaewon2@knu.ac.kr}
\authornotemark[1]
\affiliation{%
  \institution{Kyungpook National University}
  \country{Republic of Korea}
}
\orcid{0009-0001-3492-3626}

\author{Hojun Song}
\email{hojunsong@knu.ac.kr}
\affiliation{%
  \institution{Kyungpook National University}
  \country{Republic of Korea}
}
\orcid{0009-0001-3720-0415}

\author{Dongbeom Kim}
\email{kdb1209@knu.ac.kr}
\affiliation{%
  \institution{Kyungpook National University}
  \country{Republic of Korea}
}
\orcid{0009-0003-3542-9763}

\author{Junyeong Jang}
\email{wkdwnsdudas@knu.ac.kr}
\affiliation{%
  \institution{Kyungpook National University}
  \country{Republic of Korea}
}
\orcid{0009-0000-7830-7975}

\author{Aro Kim}
\email{arokim37@knu.ac.kr}
\affiliation{%
  \institution{Kyungpook National University}
  \country{Republic of Korea}
}
\orcid{0009-0001-8585-5993}

\author{Gahyeon Kim}
\email{hwang190190@knu.ac.kr}
\affiliation{%
  \institution{Kyungpook National University}
  \country{Republic of Korea}
}
\orcid{0009-0001-7798-471X}

\author{Heejung Choi}
\email{chj00@knu.ac.kr}
\affiliation{%
  \institution{Kyungpook National University}
  \country{Republic of Korea}
}
\orcid{0009-0005-7247-973X}

\author{Jehee Kim}
\email{kjhkjh626@knu.ac.kr}
\affiliation{%
  \institution{Kyungpook National University}
  \country{Republic of Korea}
}
\orcid{0009-0008-3778-241X}

\author{Gianella Cravioto}
\email{gianellacrv9@knu.ac.kr}
\affiliation{%
  \institution{Kyungpook National University}
  \country{Republic of Korea}
}
\orcid{0009-0008-2835-0076}

\author{Sohyun Lee}
\email{leeibc0414@knu.ac.kr}
\affiliation{%
  \institution{Kyungpook National University}
  \country{Republic of Korea}
}
\orcid{0009-0003-1445-3350}

\author{Gyeongjin Choi}
\email{ruen5228@knu.ac.kr}
\affiliation{%
  \institution{Kyungpook National University}
  \country{Republic of Korea}
}
\orcid{0009-0007-4648-2948}

\author{EunHye Jeong}
\email{mgs10204@knu.ac.kr}
\affiliation{%
  \institution{Kyungpook National University}
  \country{Republic of Korea}
}
\orcid{0009-0008-1599-6409}

\author{Soo Ye Kim}
\authornote{Corresponding authors}
\email{sooyek@adobe.com}
\affiliation{%
  \institution{Adobe Research}
  \country{USA}
}
\orcid{0009-0003-8104-4100}

\author{Jaehyup Lee}
\authornotemark[2]
\email{jaehyuplee@knu.ac.kr}
\affiliation{%
  \institution{Kyungpook National University}
  \country{Republic of Korea}
}
\orcid{0000-0002-2137-2650}

\author{Sang-hyo Park}
\authornotemark[2]
\email{s.park@knu.ac.kr}
\affiliation{%
  \institution{Kyungpook National University}
  \country{Republic of Korea}
}
\orcid{0000-0002-7282-7686}

\renewcommand{\shortauthors}{Kim et al.}

\begin{teaserfigure}
\centering
\includegraphics[width=\textwidth]{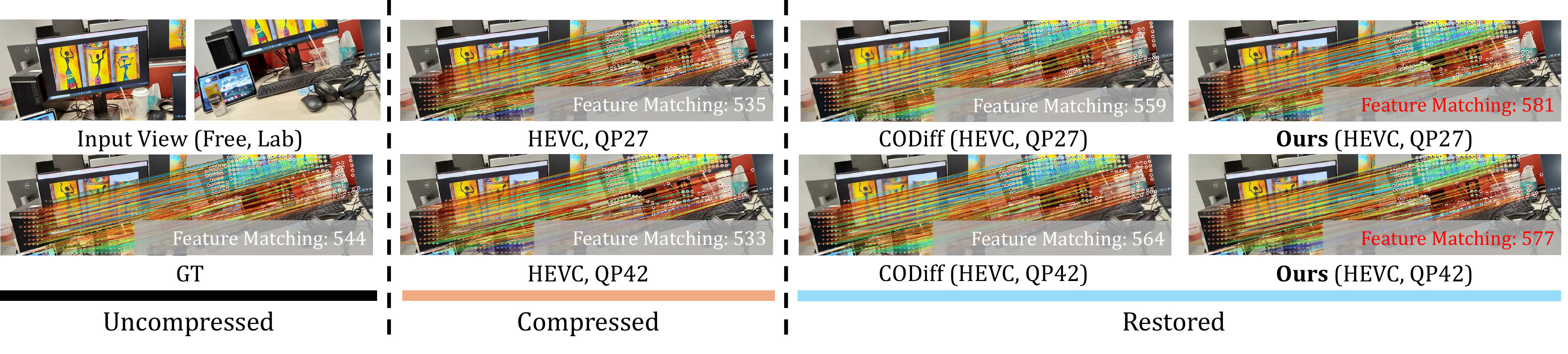}
\caption{Effect of HEVC compression on VGGT feature matching in the Free/Lab scene. Lossy compression degrades cross-view correspondence: as the quantization parameter (QP) increases from 27 to 42, the number of matched features decreases compared to the uncompressed input. While existing diffusion-based compression-aware restoration methods recover visual quality, they do not consistently restore geometry-sensitive correspondences. Our method produces more reliable feature matches under both compression levels, demonstrating improved cross-view consistency for downstream 3D reconstruction.}
\label{fig:teaser}
\end{teaserfigure}

\begin{abstract}
Real-world 3D reconstruction pipelines often rely on multi-view images degraded by lossy compression. While compression artifacts have been extensively studied in single-image or video restoration, their impact on multi-view geometry remains largely overlooked. In this work, we identify compression-induced cross-view geometric inconsistency as a critical failure mode, where view-dependent artifacts disrupt feature matching, dense correspondence, camera parameter estimation, and downstream 3D reconstruction tasks such as novel view synthesis. To address this problem, we propose \textbf{CompMVR}, a compression-aware multi-view diffusion restoration framework that jointly refines compressed views while preserving cross-view consistency, and construct compressed datasets for systematic evaluation. Unlike conventional restoration methods that optimize each image independently for 2D perceptual quality, CompMVR leverages learned compression priors and cross-view correspondence to recover visual details while improving geometric reliability across views. Experiments across diverse datasets, codecs, and compression levels demonstrate gains in restoration quality and downstream 3D tasks, including novel-view synthesis, camera pose estimation, and view matching. These results establish compressed multi-view restoration as a distinct problem and highlight its importance for 3D reconstruction under lossy compression. Code and data for this paper are available at \url{https://mchaewon.github.io/CompMVR-project-page/}.
\end{abstract}




\maketitle
\begin{table}[t]
\caption{Effect of lossy compression on VGGT view matching on RealEstate10K. We assess cross-view consistency restoration using the number of matched correspondences and AUC@5 for pose accuracy. Our method achieves the best view matching performance, outperforming diffusion-based image, video, and multi-view restoration baselines. The \colorbox{best}{Best} and \colorbox{second}{Second-best} results are highlighted in red and orange, respectively.}
\centering
\label{tab:hevc_matching}
\resizebox{\columnwidth}{!}{
\begin{tabular}{llcccc}
\hline
\multicolumn{2}{c}{Method}                  & MeanMatches $\uparrow$   & AUC@5 $\uparrow$ \\  \hline
Uncompressed    & GT                        & 2,001             & 24.55      \\ \hline
Compressed      & HEVC, QP42                & 1,655             & 22.59      \\ \hline
\multirow{4}{*}{Restored}   & CODiff$^{*}$        & 1,625             & 23.36      \\
                            & DiQP$^{*}$          & \second{1,643}    & \second{23.45}      \\
                            & SIR-Diff$^{*}$      & 1,627             & 23.09            \\
                            & Ours          & \best{1,659}      & \best{24.34}       \\
\hline
\end{tabular}
}
\begin{minipage}{\linewidth}
\footnotesize
\raggedright
$^{*}$ denotes that the model has been finetuned on our training datasets. 
\end{minipage}
\end{table}

\section{Introduction}
Modern 3D reconstruction models~\cite{kerbl20233d, wang2024dust3r, leroy2024grounding, wang2025vggt} have achieved remarkable progress, yet most do not explicitly model compression artifacts that commonly appear in real-world visual inputs. In real-world deployments, however, visual inputs are frequently stored or transmitted in compressed forms, such as JPEG~\cite{wallace1991jpeg}, H.265/HEVC~\cite{sullivan2012overview}, and H.266/VVC~\cite{bross2021overview}. Existing compression artifact restoration methods~\cite{dehaghi2025reversing} mainly formulate compression degradation at the frame level, aiming to recover high-fidelity visual appearance. Recently, latent diffusion-based models~\cite{guo2025compression, yang2026sodiff} have shown strong restoration capability by recovering image details degraded by lossy compression. Despite their effectiveness, existing methods remain largely image or video centric, without explicitly optimizing the multi-view geometric consistency required for 3D reconstruction. As shown in Fig.~\ref{fig:teaser}, restoration can substantially affect the reliability of cross-view feature matching. However, this geometric aspect has been largely overlooked in prior restoration methods.

Recent diffusion-based restoration methods~\cite{lin2024diffbir, wu2024one, yu2024scaling, guo2025compression, yang2026sodiff} have demonstrated strong generative priors for recovering perceptually realistic image details. They have been successfully explored for blind image restoration~\cite{lin2024diffbir}, real-world image super-resolution~\cite{wu2024one}, photo-realistic image restoration~\cite{yu2024scaling}, and compression artifact removal~\cite{guo2025compression, yang2026sodiff}. However, these methods are primarily formulated as image or video restoration problems, where the goal is to improve visual fidelity, perceptual quality, or artifact reduction. When such methods are applied to compressed multi-view observations, restoration is typically performed in a view-wise manner. As a result, they do not explicitly enforce cross-view geometric consistency (see Tab.~\ref{tab:hevc_matching}), which is essential for reliable correspondence estimation and downstream 3D reconstruction.

Motivated by these limitations, we newly propose \textbf{CompMVR}, a compression-aware multi-view restoration diffusion framework designed to improve geometric reliability for downstream 3D reconstruction. Our key observation is that compressed multi-view inputs can exhibit compression-induced inconsistency across views: artifacts depend on both the codec and QP, and thus may unsettle geometry-sensitive cues differently across views. To analyze this failure systematically, we construct a compressed multi-view benchmark by synthesizing compressed versions of existing 3D datasets under multiple codecs and quantization levels, while preserving codec and QP annotations. CompMVR uses codec and QP supervision during training, while inferring compression cues directly from the input at inference. Our benchmark shows that compression can substantially degrade geometry-sensitive tasks, including novel view synthesis, view matching, and camera pose estimation.

To address the challenge of handling view-dependent and spatially varying compression artifacts, CompMVR adopts a two-stage restoration framework for compressed multi-view inputs. In the first stage, CompMVR derives coding priors under codec and QP supervision during training via a compression prior embedder, explicitly characterizing the degradation regime for subsequent diffusion-based restoration. The coding priors provide the global compression context. In the second stage, the model introduces spatially varying artifact cues that capture the local reliability of geometry-critical regions, and jointly refines multiple views guided by these cues. To explicitly enforce geometric consistency, we incorporate correspondence-based supervision that aligns restoration across geometrically corresponding regions, while leveraging multi-view 3D attention to facilitate feature interaction across views. CompMVR combines compression-aware conditioning, multi-view interaction, and correspondence supervision for geometric consistency. The key contributions of our approach are as follows:
\begin{itemize}
  \item We newly formulate compression-induced cross-view geometric inconsistency and introduce a codec-synthesized multi-view benchmark with clean/compressed pairs, codec labels, and QP annotations. 
  \item We propose CompMVR, a compression-aware multi-view diffusion framework that integrates codec- and QP-conditioned global priors with spatially varying artifact estimation and correspondence-based supervision to improve cross-view geometric reliability during restoration.
  \item We show consistent gains across image restoration metrics and downstream 2D/3D vision tasks, including feature matching, camera parameter estimation, point map prediction, novel-view synthesis, and 2D quality assessment with generalization across codecs and compression levels.
\end{itemize}
\section{Related work}
\begin{figure*}[t]
\centering
\includegraphics[width=0.92\textwidth]{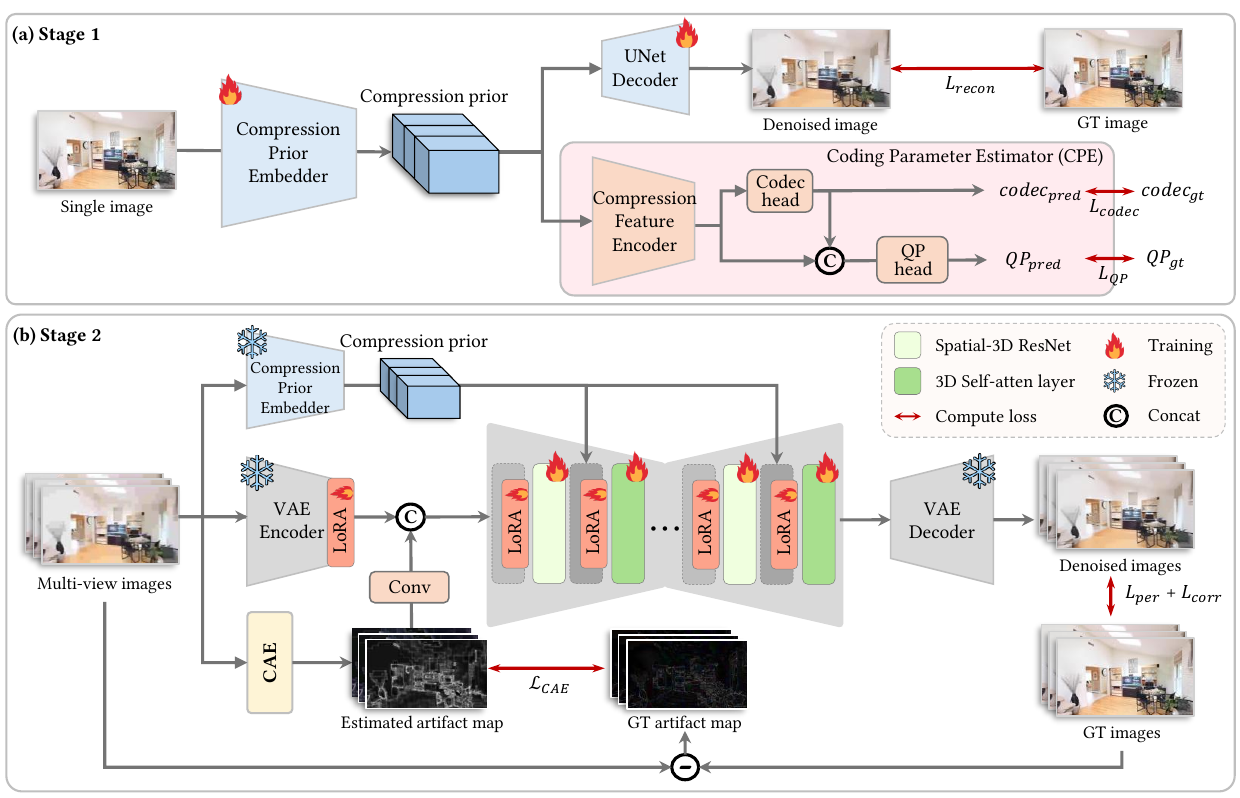}
\caption{Overview of our proposed CompMVR. (a) In Stage 1, we learn a compression-aware latent representation by training a compression prior embedder. The latent is optimized through image reconstruction and supervision of coding parameters (codec type and QP). 
(b) In Stage 2, we perform multi-view restoration using a diffusion model conditioned on the learned compression prior. The latent is injected via cross-attention and fused with the UNet input, while a compression artifact estimator (CAE) predicts spatially varying residual maps for degradation-aware restoration. To enforce cross-view geometric consistency, we further apply a cross-view correspondence loss during training.}
\label{fig:main_framework}
\end{figure*}

\subsection{Compression-Aware Restoration}
Recently, diffusion-based restoration methods~\cite{welker2024driftrec, guo2025compression, yang2026sodiff} have shown notable performance in compression artifact removal. Compared with transformer-based models~\cite{liang2021swinir, zamir2022restormer, chen2022cross}, diffusion models provide generative priors through iterative denoising. CPGA~\cite{zhu2024cpga} pioneers the use of explicit coding priors, such as motion vectors, residual frames, and predictive frames. More recent studies further incorporate degradation-related cues, such as quality factors, into diffusion-based restoration~\cite{guo2025compression, yang2026sodiff}. These approaches demonstrate that explicit coding information can provide useful guidance for compression artifact removal. However, compression-aware restoration methods remain primarily focused on visual enhancement. Consequently, they do not explicitly optimize cross-view geometric reliability for downstream 3D reconstruction. 
Therefore, existing works remain largely geometry-agnostic, as they do not explicitly adapt restoration to the cross-view geometric requirements of downstream 3D reconstruction. In contrast, our work integrates explicit coding priors with cross-view diffusion restoration to recover geometry-consistent visual evidence from compressed multi-view inputs.

\subsection{Generative priors for 3D Reconstruction}
Diffusion-based models have been widely investigated for 3D reconstruction~\cite{liu2023zero, poole2023dreamfusion, lin2023magic3d, wang2023prolificdreamer, shi2024mvdream, long2024wonder3d, gao2024cat3d}. Recently, rather than training generative models from scratch, many approaches adapt large-scale pretrained diffusion backbones, such as Stable Diffusion~\cite{rombach2022high}, to provide strong visual priors for 3D-related tasks. In image-to-3D reconstruction, diffusion priors are often used to synthesize unseen viewpoints or enrich missing visual details under sparse or ambiguous observations~\cite{liu2023zero, poole2023dreamfusion, lin2023magic3d, wang2023prolificdreamer}. Several multi-view diffusion methods further extend image diffusion architectures by generating or refining multiple views jointly, often introducing 3D-aware attention mechanisms to encourage view consistency~\cite{shi2024mvdream, long2024wonder3d, gao2024cat3d}. These designs demonstrate that diffusion priors can provide useful structural and appearance cues when observations are incomplete or constrained. However, most existing 3D reconstruction methods do not explicitly consider degradation types. To preserve cross-view consistency, SIR-Diff~\cite{mao2025sir} states that independent single-view restoration can compromise geometric constraints and processes multiple views jointly with a 3D self-attention transformer. However, SIR-Diff is not designed for compression-aware restoration, as it does not exploit codec- or QP-dependent coding priors that characterize compression artifacts. In contrast, our work studies diffusion priors under practical compression degradation and aims to recover geometry-consistent multi-view evidence for robust 3D reconstruction.

\subsection{3D reconstruction under Degraded Inputs}
Recent 3D reconstruction methods have explored novel-view synthesis under degraded observations~\cite{zhou2024layout, huang2023refsr, feng2024srgs, zhou2023nerflix, pearl2022nan, zhou2023nerflixpp}. DIFIX3D+~\cite{wu2025difix3d+} uses diffusion priors to enhance artifact-contaminated NeRF/3DGS renderings, while HQGS~\cite{lin2025hqgs} studies robust 3DGS under degraded observations and CLIPGaussian~\cite{cao2025enhancing} incorporates semantic guidance and 3D spatial consistency for low-quality inputs. Other works address specific degradations such as extreme motion blur to improve camera estimation, geometry recovery, and reconstruction~\cite{jang2025splat}. Collectively, these studies show that degradation affects both visual quality and 3D geometry. However, they focus primarily on generic low-quality inputs, rendering artifacts, or specific capture degradations rather than codec-dependent compression in multi-view observations. In contrast, CompMVR explicitly models input-side lossy compression with compression-aware priors and cross-view supervision to improve geometric reliability before downstream 3D reconstruction.

\section{Proposed Method}
\subsection{Stage 1: Compression-Aware Representation Learning}
An overview of our method is shown in Fig.~\ref{fig:main_framework}. Stage 1 aims to learn a compression-aware latent representation that captures degradation characteristics induced by lossy codecs, as illustrated in Fig.~\ref{fig:main_framework}(a). 
Given a compressed input image $I_c \in \mathbb{R}^{H\times W \times 3}$, the encoder produces a compression-aware latent representation learned through compression parameter prediction. To regularize the learning of this representation, the latent is additionally used for image reconstruction via a decoder.

We introduce a coding parameter estimator (CPE) to provide supervision for learning the latent representation. The CPE first extracts a shared feature using a compression feature encoder.
The codec head then predicts the codec type, and a soft codec embedding is computed based on the predicted class probabilities. 
To model the dependency between codec type and QP, the soft codec embedding is concatenated with the shared feature and fed into the QP prediction head. This design encourages the latent representation to capture codec-dependent degradation characteristics.

The QP head is supervised with an L1 regression loss:
\begin{equation}
\mathcal{L}_{\text{QP}} = \| \hat{QP} - QP_{\text{gt}} \|_1,
\end{equation}
and the codec head is supervised with a cross-entropy classification loss:
\begin{equation}
\mathcal{L}_{\text{codec}} = -\sum_{c} \mathbf{1}[c = c_{\text{gt}}] \log \hat{p}_c.
\end{equation}
The coding parameter estimation objective jointly optimizes both losses as 
$
\mathcal{L}_{\text{CPE}} = \lambda_{\text{QP}}\mathcal{L}_{\text{QP}} + \lambda_{\text{codec}} \mathcal{L}_{\text{codec}}.
$
The final Stage 1 loss $\mathcal{L}_{\text{CaRL}}$ combines the reconstruction loss and the compression-aware losses:
\begin{equation}
\mathcal{L}_{\text{CaRL}} = \| I_{recon} - I_{\text{gt}} \|_1 + \mathcal{L}_{\text{CPE}}.
\end{equation}
As a result, the learned latent representation encodes global codec- and QP-dependent degradation context. Codec and QP labels are used only for training, while the compression prior is inferred from the input at inference.

\subsection{Stage 2: Multi-View Restoration Diffusion Model}
As shown in Fig.~\ref{fig:main_framework}(b), we train a restoration diffusion model that jointly processes $N$ compressed images $\{\mathbf{I}^i_c\}^N_{i=1}$. In our implementation, we utilize $N=4$. The UNet backbone is extended with modules for spatially-aware degradation modeling and geometry-aware cross-view consistency.

\paragraph{Spatially-variant Compression Artifact Estimation}
\begin{figure}
  \centering
  \includegraphics[width=0.90\columnwidth]{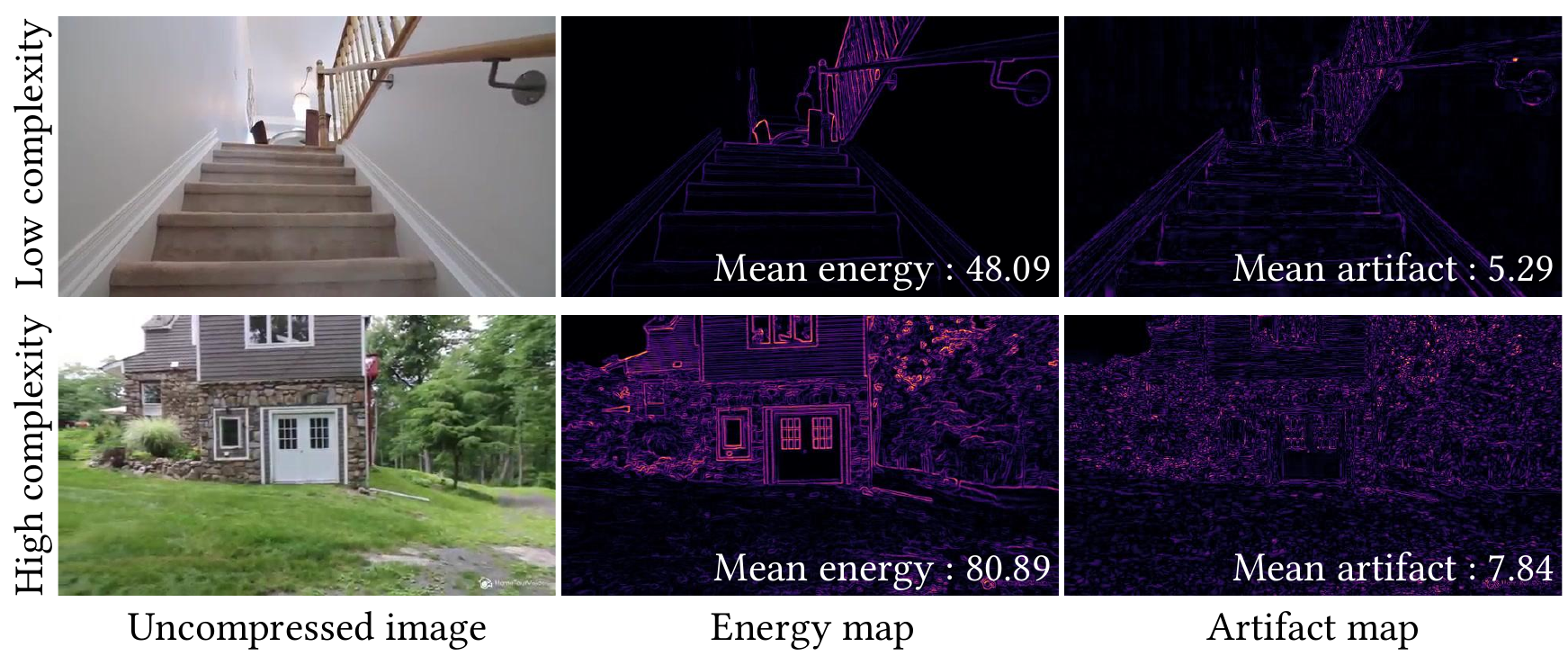}
  \caption{Energy maps computed from uncompressed images and artifact maps compressed at identical QP settings, for low and high complexity scenes. Higher scene complexity yields larger compression residuals.}
  \label{fig:cre_motivation}
\end{figure}

We explicitly model spatially varying, content-dependent degradation within each view. While Stage 1 captures global compression characteristics such as codec type and QP, these signals cannot fully describe local artifact variations. To quantify this, we use a gradient-based energy function, $e(I)=|\partial_x I|+|\partial_y I|$, defined as the $L1$ norm of image gradients and commonly used in seam carving~\cite{avidan2023seam}. Low-energy regions correspond to smooth areas, whereas high-energy regions capture texture-rich or edge-dominant structures. As shown in Fig.~\ref{fig:cre_motivation}, the energy map strongly correlates with the artifact map, with high-complexity regions exhibiting larger residuals than smooth regions even under identical QP settings. This shows that compression distortion depends not only on global parameters but also on local image structure, motivating explicit spatial degradation modeling.

To address this, we introduce a compression artifact estimator (CAE), inspired by CBDNet~\cite{guo2019cbdnet}, to model spatially varying degradation from compressed inputs. CAE is a lightweight 11-layer fully convolutional network with 0.375M parameters that predicts
$
\hat{r}(I_{c}) = F_{\text{CAE}}(I_{c};\,\mathcal{W})
$
directly from the compressed image, without requiring a high-quality reference at inference. Its encoder-decoder architecture with symmetric skip connections captures both local and global context, highlighting compression-sensitive regions such as edges, fine textures, and high-frequency structures.

The estimated artifact map is used as a spatial conditioning signal for restoration. Specifically, it is projected into the latent space and concatenated with the encoded latent $\mathbf{z}_L$:
\begin{equation}
\mathbf{z}' = \text{Concat}(\mathbf{z}_L,\, \phi(\hat{r})),
\end{equation}
where $\phi(\cdot)$ projects the CAE output to the spatial resolution and channel dimension of the diffusion latent before concatenation. The augmented latent $\mathbf{z}'$ is then fed into the UNet, enabling spatially-adaptive restoration.

We train the CAE using the ground-truth residual map $r_{\text{gt}} = |I_{gt} - I_{c}|,$ where $I_{gt}$ denotes the uncompressed images. This serves as a spatial supervision signal indicating compression-sensitive regions.
Since under-estimation may cause the restoration network to apply insufficient correction, we adopt an asymmetric loss:
\begin{equation}
\mathcal{L}_{\text{asymm}} = \frac{1}{N}\sum_i \bigl|\alpha - \mathbf{1}[\epsilon_i < 0]\bigr| \cdot \epsilon_i^2,
\end{equation}
where $\epsilon = \hat{r} - \tilde{r}_{\text{gt}}$, $\tilde{r}_{\text{gt}} = r_{\text{gt}}/2$, and $\alpha = 0.3$.
We further introduce a cosine similarity loss to encourage spatial alignment between the predicted and ground-truth artifact maps:
\begin{equation}
\mathcal{L}_{\text{spatial}} = 1 - \frac{\hat{\mathbf{r}} \cdot \tilde{\mathbf{r}}_{\text{gt}}}{\|\hat{\mathbf{r}}\|_2 \|\tilde{\mathbf{r}}_{\text{gt}}\|_2}.
\end{equation}
The final training objective for CAE is defined as: 
\begin{equation}
\mathcal{L}_{\text{CAE}} = \mathcal{L}_{\text{asymm}} + \mathcal{L}_{\text{spatial}}.
\end{equation}

\begin{figure}[t] 
  \centering
  \includegraphics[width=\columnwidth]{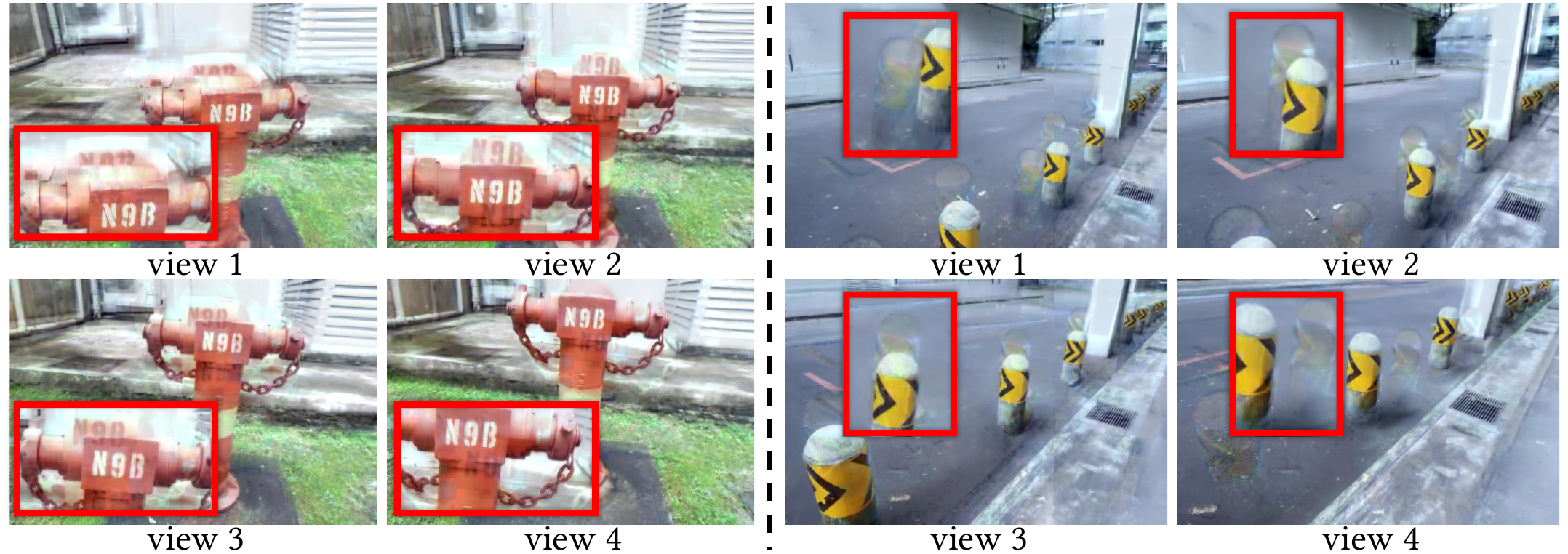}
  \caption{Failure cases of a video diffusion restoration model~\cite{dehaghi2025reversing} on multi-view images. Examples from the Free dataset (hydrant and pillar scenes, left to right) show that the model fails to maintain cross-view consistency. Highlighted regions reveal inconsistent restoration and geometric misalignment caused by large viewpoint differences.}
  \label{fig:multiview_inconsistency}
\end{figure}

\begin{figure*}[p]
  \centering
  \includegraphics[width=\textwidth]{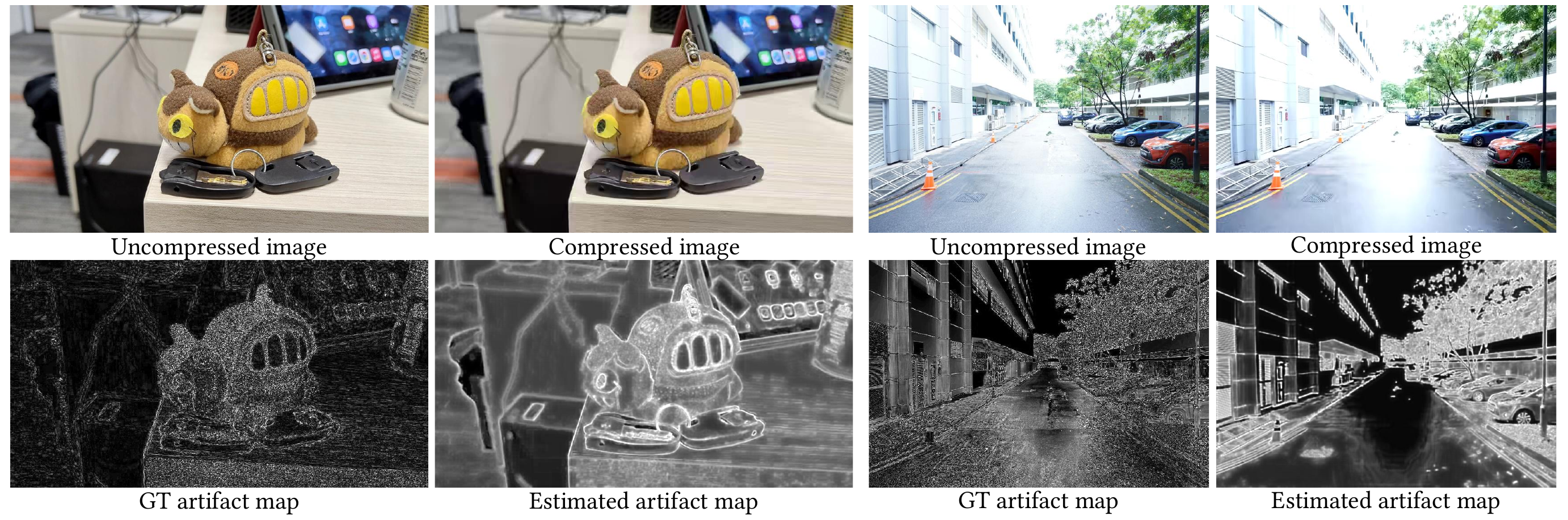}
  \caption{Visualization of estimated artifact maps on the Free dataset. The estimated artifact maps highlight compression-sensitive regions, such as edges and texture-rich structures, demonstrating that our CAE effectively captures spatially varying degradation cues.}
  \label{fig:artifact_map}
\end{figure*}

Fig.~\ref{fig:artifact_map} shows qualitative comparisons between ground-truth and estimated artifact maps. These results align with the goal of CAE, which is to provide spatial degradation cues for guiding restoration rather than reconstructing exact residual values.

\begin{table*}[!htbp]
\begin{minipage}{\linewidth}
\footnotesize
\raggedright
\end{minipage}
\centering
\caption{Quantitative novel-view synthesis results using 3D Gaussian Splatting on the Free dataset under different codec and QP settings. We report PSNR, SSIM, and LPIPS, where higher PSNR/SSIM and lower LPIPS indicate better rendering quality. The \colorbox{best}{Best} and \colorbox{second}{Second-best} results are highlighted in red and orange, respectively.}
\label{tab:3dgs_results}
\resizebox{\linewidth}{!}{%
\begin{tabular}{llccccccccccccccc}
\toprule
\multirow{2}{*}{Codec} & \multirow{2}{*}{QP}
& \multicolumn{5}{c}{PSNR $\uparrow$}
& \multicolumn{5}{c}{SSIM $\uparrow$}
& \multicolumn{5}{c}{LPIPS $\downarrow$} \\
\cmidrule(lr){3-7} \cmidrule(lr){8-12} \cmidrule(lr){13-17}
&
& Compressed & CODiff$^{*}$ & DiQP$^{*}$ & SIR-Diff$^{*}$ & \textbf{Ours}
& Compressed & CODiff$^{*}$ & DiQP$^{*}$ & SIR-Diff$^{*}$ & \textbf{Ours}
& Compressed & CODiff$^{*}$ & DiQP$^{*}$ & SIR-Diff$^{*}$ & \textbf{Ours} \\
\midrule
\multirow{2}{*}{AVC}
& 27
& 22.35 & \second{20.83} & 20.50 & 19.72 & \best{22.90}
& 0.601 & 0.541 & \second{0.625} & 0.512 & \best{0.642}
& 0.309 & \second{0.378} & 0.393 & 0.426 & \best{0.236} \\
& 42
& 22.01 & \second{21.39} & 19.95 & 19.61 & \best{23.21}
& 0.644 & \second{0.586} & 0.571 & 0.517 & \best{0.688}
& 0.382 & \second{0.433} & 0.472 & 0.462 & \best{0.325} \\
\midrule
\multirow{2}{*}{HEVC}
& 27
& 23.45 & \second{22.11} & 20.54 & 20.07 & \best{24.08}
& 0.638 & 0.580 & \second{0.607} & 0.516 & \best{0.670}
& 0.300 & \second{0.382} & 0.425 & 0.447 & \best{0.242} \\
& 42
& 22.96 & \second{22.49} & 21.10 & 19.90 & \best{24.49}
& 0.687 & \second{0.605} & 0.595 & 0.517 & \best{0.718}
& 0.285 & \second{0.403} & 0.451 & 0.464 & \best{0.265} \\
\midrule
\multirow{2}{*}{AV1}
& 27 %
& 22.37 & \second{20.87} & 20.27 & 19.63 & \best{22.84}
& 0.605 & \second{0.550} & 0.404 & 0.512 & \best{0.647}
& 0.321 & \second{0.379} & 0.618 & 0.431 & \best{0.238} \\
& 42 %
& 22.61 & \second{21.27} & 20.26 & 19.40 & \best{23.49}
& 0.696 & \second{0.591} & 0.602 & 0.513 & \best{0.701}
& 0.315 & \second{0.401} & 0.445 & 0.454 & \best{0.270} \\
\midrule
\multirow{2}{*}{VP9}
& 27
& 22.31 & \second{20.84} & 20.40 & 19.68 & \best{22.85}
& 0.629 & 0.541 & \second{0.628} & 0.511 & \best{0.638}
& 0.315 & \second{0.377} & 0.387 & 0.425 & \best{0.235} \\
& 42 %
& 22.49 & \second{21.24} & 20.56 & 19.76 & \best{23.30}
& 0.651 & \second{0.580} & 0.616 & 0.518 & \best{0.685}
& 0.292 & \second{0.396} & 0.423 & 0.441 & \best{0.268} \\

\midrule
  \multicolumn{2}{l}{Uncompressed}
  & \multicolumn{5}{c}{28.30}
  & \multicolumn{5}{c}{0.885}
  & \multicolumn{5}{c}{0.112} \\
\bottomrule
\end{tabular}}
\begin{minipage}{\linewidth}
\footnotesize
\raggedright
$^{*}$ denotes that the model has been finetuned on our training datasets. 
\end{minipage}
\end{table*}

\subsubsection{Inter-View Geometric Consistency}
Maintaining consistency across views is crucial for geometrically coherent restoration. Unlike video diffusion models designed for sequential frames with small viewpoint changes~\cite{dehaghi2025reversing}, multi-view images exhibit large viewpoint disparities, where corresponding regions are geometrically displaced. As shown in Fig.~\ref{fig:multiview_inconsistency}, applying a video diffusion model out of the box leads to view-inconsistent artifacts. This observation clearly demonstrates the necessity of geometry-aware consistency modeling for multi-view restoration.
This challenge arises from the difficulty of establishing reliable correspondences under large viewpoint changes, occlusions, and view-dependent artifacts. Therefore, a robust mechanism for estimating 
geometry-aware correspondences is essential for enforcing cross-view consistency.

\paragraph{Cross-View Correspondence Loss}
To address this challenge, we introduce a cross-view correspondence loss based on a pre-trained VGGT~\cite{wang2025vggt}, which provides per-pixel world coordinates and confidence maps for training-time supervision. VGGT is not required during restoration inference. 
Correspondences are computed from high-quality reference images via 3D nearest-neighbor matching in world-coordinate space, followed by confidence and geometric filtering to retain reliable matches.
The correspondence loss is defined as a confidence-weighted photometric loss over the filtered correspondences:
\begin{equation}
\mathcal{L}_{\text{corr}} =
\frac{1}{|\mathcal{P}|}
\sum_{(i,j)\in\mathcal{P}}
\frac{
\sum_{p \in \mathcal{M}_{i \rightarrow j}}
w_{pq^*}
\left\|\hat{\mathbf{{I}}}_i^p - \mathbf{I}_{gt,j}^{q^*}\right\|_1
}{
\sum_{p \in \mathcal{M}_{i \rightarrow j}} w_{pq^*}
},
\label{eq:l_corr}
\end{equation}
where $\mathbf{\hat{{I}}}_i$ is the restored image of the input $i$, \(\mathcal{P}\) denotes the set of sampled view pairs, \(\mathcal{M}_{i \rightarrow j}\) is the set of reliable 
correspondences, \(q^*\) is the matched pixel in view \(j\), and \(w_{pq^*}\) is a confidence weight derived from predictions of VGGT. When severe compression produces unreliable correspondences, low-confidence regions are down-weighted, reducing their contribution to $\mathcal{L}_{\text{corr}}$. The model therefore relies more on reconstruction loss, the Stage-1 prior, and CAE rather than enforcing unreliable correspondences.  %This loss enforces that restored images are consistent with geometrically corresponding regions across views, thereby improving multi-view consistency and stabilizing downstream 3D reconstruction.

\paragraph{Implicit Cross-View Consistency via 3D Attention}
We further incorporate Spatial-3D ResNet~\cite{mao2025sir, shi2024mvdream} and 3D self-attention layers~\cite{mao2025sir} to model cross-view interactions, following prior multi-view diffusion approaches. Unlike SIR-Diff, which initializes the 3D modules from pre-trained Stable Video Diffusion (SVD) weights, we train all 3D components from scratch using zero initialization to avoid introducing temporal continuity priors that are not well aligned with multi-view image settings. These 3D layers enable implicit cross-view interaction by jointly modeling multi-view features. However, they do not explicitly model view-dependent degradation, as the attention mechanism is driven by feature similarity without an explicit notion of per-view quality. Our method complements this implicit modeling with explicit correspondence-based supervision in Eq.~\ref{eq:l_corr}, resulting in more robust multi-view consistency under view-varying compression artifacts.

\paragraph{Overall Objective}
% $\mathcal{L}_{\text{CAE}}$ supervises the compression artifact estimator to capture spatially varying degradation, while $\mathcal{L}_{\text{corr}}$ enforces cross-view geometric consistency through reliable correspondences. 
To further improve perceptual quality, we incorporate a perceptual reconstruction loss defined as:
\begin{equation}
\mathcal{L}_{\text{per}} = \mathcal{L}_{\text{recon}} + \mathcal{L}_{\text{DISTS}},
\end{equation}
where $\mathcal{L}_{\text{recon}}$ is a pixel-wise reconstruction loss and $\mathcal{L}_{\text{DISTS}}$ measures perceptual similarity between the restored and ground-truth images. 
The overall objective in Stage 2 is jointly defined as:
\begin{equation}
\mathcal{L} = \mathcal{L}_{\text{CAE}} + \mathcal{L}_{\text{corr}} + \mathcal{L}_{\text{per}}.
\end{equation}
This formulation enables the model to simultaneously handle spatially varying compression artifacts, enforce multi-view geometric consistency, and preserve perceptual fidelity.

\section{Experiments}
We train Stage 1 for 200K iterations with a batch size of 8, and Stage 2 for 100K iterations with a batch size of 2 and 4 views, both on a single NVIDIA A6000 GPU. We adopt the Stable Diffusion 2.1 backbone with LoRA~\cite{hu2022lora} fine-tuning.

\subsection{Datasets}
We train our model on 146 scenes from DTU~\cite{jensen2014large}, LLFF~\cite{mildenhall2019local}, Mip-NeRF 360~\cite{barron2022mip}, and Tanks and Temples~\cite{knapitsch2017tanks}, using 140 for training and 6 for validation. For evaluation, we use Free~\cite{wang2023f2}, CO3Dv2~\cite{reizenstein2021common}, and RealEstate10K~\cite{zhou2018stereo} to cover diverse 3D reconstruction scenarios. All scenes are encoded using the compression settings provided in the supplementary material. Both training and test datasets use four codecs: H.264/AVC~\cite{sullivan2004h}, H.265/HEVC~\cite{sullivan2012overview}, VP9~\cite{mukherjee2015technical}, and AV1~\cite{chen2020overview}. For training, we adopt six quantization levels: QP 20, 26, 32, 38, 44, and 50. For testing, we use two quantization levels: QP 27 and 42, both unseen during training. Since QP values are codec-specific, the same numerical QP does not necessarily represent equivalent degradation severity across codecs. Detailed encoding parameters, scene lists, and compression configurations are provided in the supplementary material.

\subsection{3D Evaluation}
\begin{figure*}[p]
    \centering
    \includegraphics[width=0.95\textwidth]{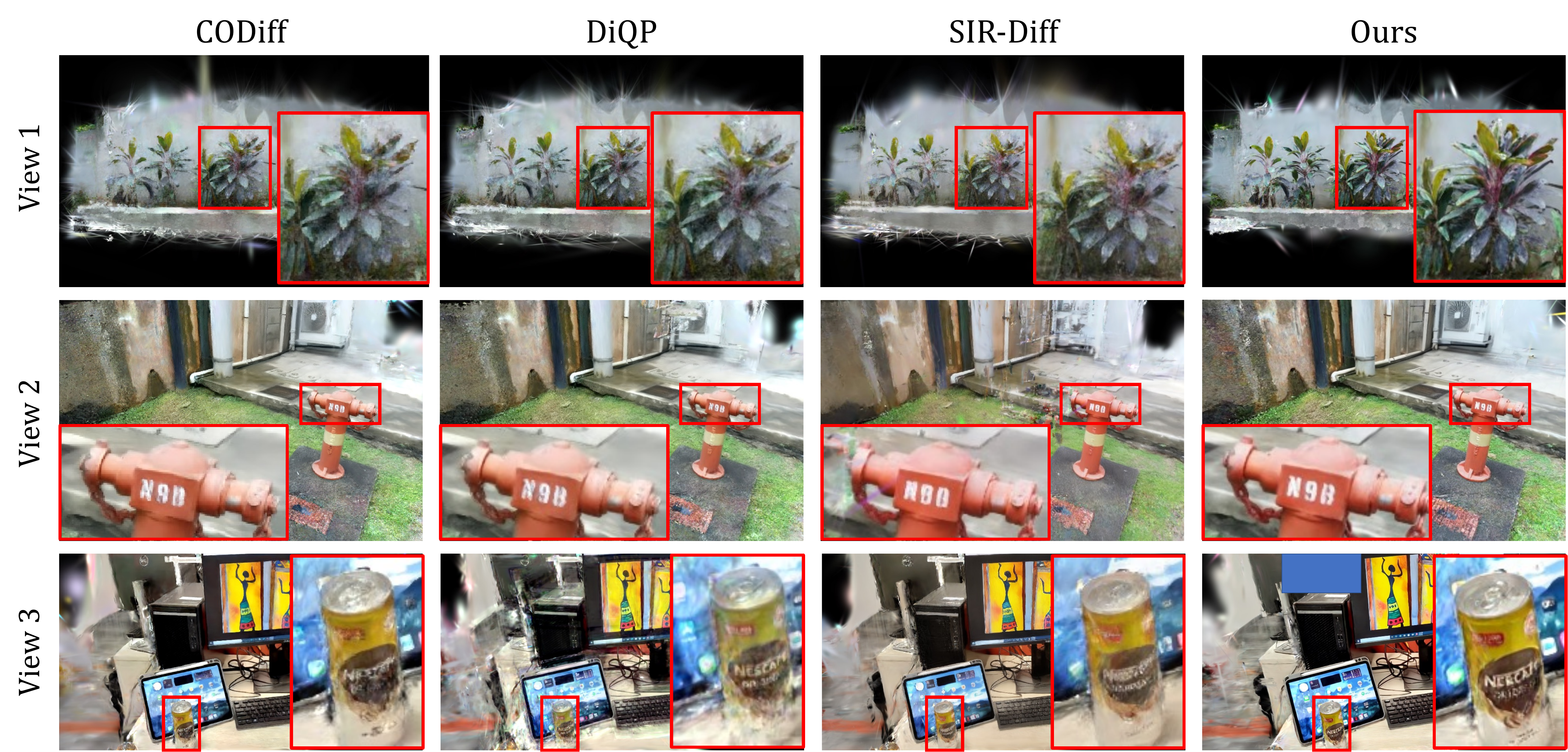}
    \caption{Qualitative comparison of synthesized novel 3D views from restored multi-view inputs on the Free dataset. We evaluate global consistency using scene renderings synthesized from multiple viewpoints and local structural fidelity using zoomed-in regions. Views 1 and 2 correspond to HEVC QP42, and View 3 corresponds to AV1 QP42. Our method preserves sharper geometry-critical details and reduces blur and distortion compared with diffusion-based restoration baselines. Red boxes indicate representative regions for comparison.}
    \label{fig:3D_global}
\end{figure*}

\begin{figure*}[p]
    \centering
    \includegraphics[width=0.90\textwidth]{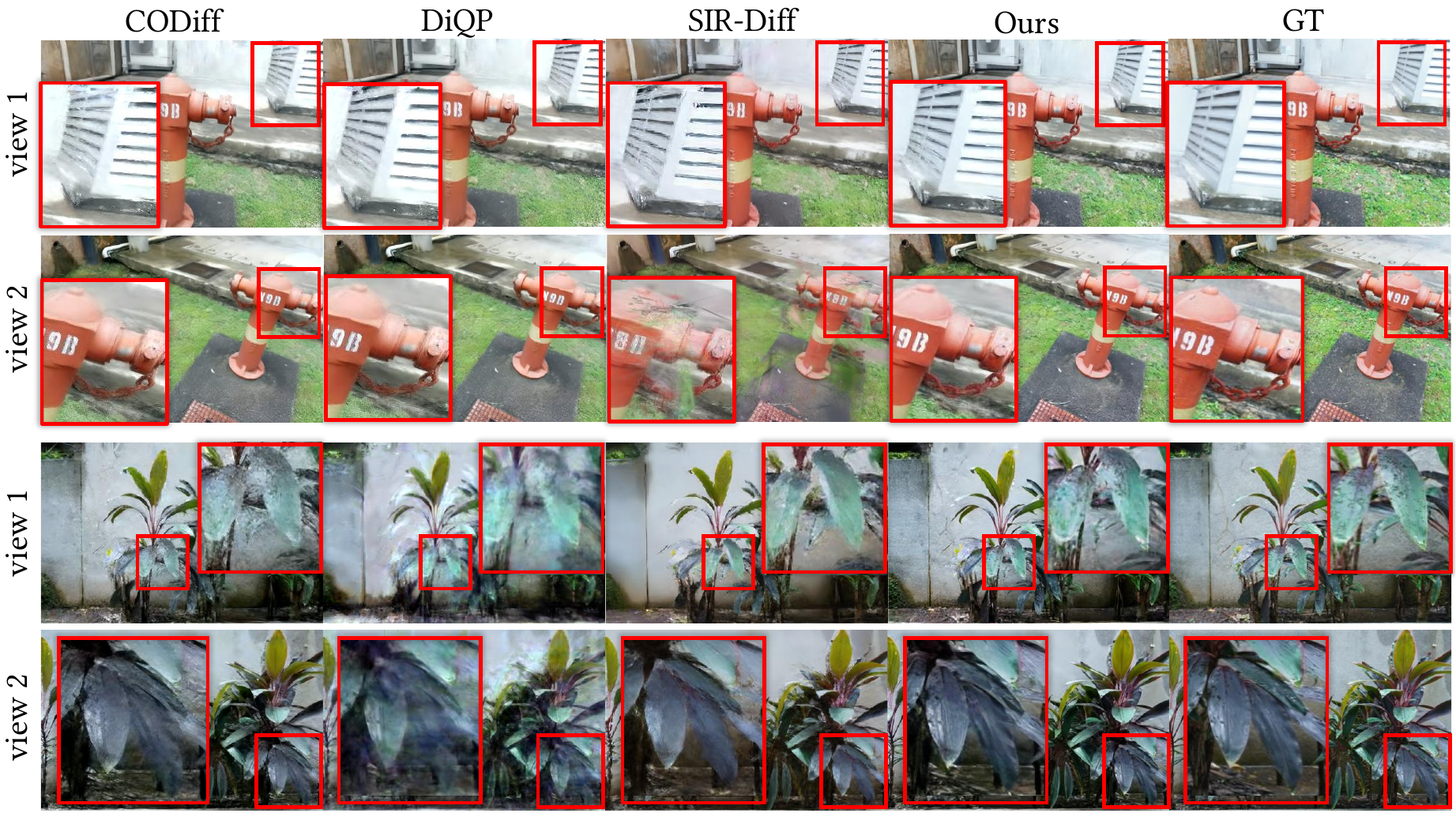}
    \caption{Qualitative novel-view synthesis results rendered at ground-truth target camera poses. The top two rows correspond to HEVC QP42, and the bottom two rows correspond to AVC QP42. We compare each method with the corresponding GT view using full-scene renderings and zoomed-in regions. Our method better preserves object structures, text regions, and high-frequency details, producing results closer to GT than diffusion-based restoration baselines. Red boxes indicate representative regions for comparison.}
    \label{fig:3D_local}
\end{figure*}

\paragraph{Novel-view Synthesis.} For the evaluation of novel-view synthesis, we train a 3D Gaussian Splatting model~\cite{kerbl20233d} on restored multi-view inputs from the Free dataset. PSNR, SSIM, and LPIPS are reported in Tab.~\ref{tab:3dgs_results}. As shown in Tab.~\ref{tab:3dgs_results}, our method consistently achieves the best performance across all codec and QP settings. Compared with the second-best baseline, our method improves PSNR/SSIM by 2.02 dB/0.071 and reduces LPIPS by 0.134, demonstrating more reliable inputs for downstream 3DGS. As a result, Figs.~\ref{fig:3D_global} and~\ref{fig:3D_local} show that our method produces more coherent novel-view renderings and sharper local structures than the baselines. In particular, our results better preserve fine object boundaries, text regions, and high-frequency details across views.

\paragraph{Geometric Evaluation} We evaluate downstream geometry using VGGT~\cite{wang2025vggt}. Camera pose estimation and view matching are evaluated on RealEstate10K using AUC@15/30 and AUC@5, respectively (Tabs.~\ref{tab:camera_pose_estimation} and~\ref{tab:view_matching}), while point map estimation is evaluated on CO3Dv2 and reported in the supplementary material. Our method achieves the best AUC@15/30 across all codecs and compression levels and the highest average AUC@5, indicating more reliable geometry and cross-view correspondences under compression. We further evaluate multi-view consistency using MEt3R~\cite{asim2025met3r} under matched fine-tuning (Tab.~\ref{tab:matched_finetune}), where ours achieves the lowest score of 0.132. SIR-Diff achieves a lower point-map Chamfer distance; additional protocols and results are provided in the supplementary material.

\begin{table}[!tbp]
\centering
\setlength{\tabcolsep}{2.5pt}
\renewcommand{\arraystretch}{0.92}
\caption{VGGT camera pose estimation results on the RealEstate10K dataset under different codec and QP settings. AUC@15 and AUC@30 are reported, where higher values indicate better pose estimation accuracy. The \colorbox{best}{Best} and \colorbox{second}{Second-best} results are highlighted in red and orange, respectively.}
\label{tab:camera_pose_estimation}
\resizebox{\linewidth}{!}{%
\begin{tabular}{llcccccccccc}
\toprule
\multirow{2}{*}{Codec} & \multirow{2}{*}{QP} 
& \multicolumn{5}{c}{AUC@15 $\uparrow$} 
& \multicolumn{5}{c}{AUC@30 $\uparrow$} \\
\cmidrule(lr){3-7} \cmidrule(lr){8-12}
& 
& Compressed & CODiff$^{*}$ & DiQP$^{*}$ & SIR-Diff$^{*}$ & \textbf{Ours}
& Compressed & CODiff$^{*}$ & DiQP$^{*}$ & SIR-Diff$^{*}$ & \textbf{Ours} \\
\midrule
\multirow{2}{*}{AVC} 
& 27 & 64.72 & 72.27 & 70.05 & \second{73.28} & \best{77.52} & 77.36 & 81.04 & 81.30 & \second{81.71} & \best{85.00} \\
& 42 & 61.37 & 69.36 & 66.16 & \second{69.50} & \best{73.22} & 76.20 & 79.02 & 78.69 & \second{79.26} & \best{82.79} \\ 
\midrule
\multirow{2}{*}{HEVC}
& 27 & 64.12 & 72.91 & 70.11 & \second{73.35} & \best{79.32} & 76.84 & \second{81.38} & 81.33 & 81.26 & \best{87.52} \\
& 42 & 63.77 & \second{71.44} & 69.70 & 69.61 & \best{75.26} & 76.00 & 79.82 & \second{80.75} & 79.07 & \best{83.70} \\
\midrule
\multirow{2}{*}{VP9} 
& 27 & 61.90 & \second{73.14} & 70.85 & 72.65 & \best{77.56} & 75.38 & 81.45 & \second{81.95} & 80.92 & \best{85.32} \\
& 42 & 58.27 & \second{71.87} & 69.28 & 71.42 & \best{75.80} & 73.33 & 80.36 & \second{80.73} & 80.00 & \best{83.99} \\
\midrule
\multirow{2}{*}{AV1} 
& 27 & 66.54 & 73.03 & 71.97 & \second{73.64} & \best{78.58} & 78.84 & 81.40 & \second{82.39} & 81.43 & \best{86.53} \\
& 42 & 62.59 & \second{73.48} & 71.48 & 73.28 & \best{77.88} & 75.98 & 81.73 & \second{81.75} & 81.59 & \best{85.93} \\
\bottomrule
\end{tabular}
}
\begin{minipage}{\linewidth}
\footnotesize
\raggedright
$^{*}$ denotes that the model has been finetuned on our training datasets. 
\end{minipage}
\centering
\end{table}

\begin{table}[!htbp]
\centering
\caption{
VGGT view matching results on the RealEstate10K dataset under different codec and QP settings. We report AUC@5, where higher values indicate better pose estimation accuracy. The \colorbox{best}{Best} and \colorbox{second}{Second-best} results are highlighted in red and orange, respectively.
}
\label{tab:view_matching}
\setlength{\tabcolsep}{3pt}
\renewcommand{\arraystretch}{0.90}
\resizebox{0.80\columnwidth}{!}{%
\begin{tabular}{llccccc}
\toprule
\multirow{2}{*}{Codec} & \multirow{2}{*}{QP}
& \multicolumn{5}{c}{AUC@5 $\uparrow$} \\
\cmidrule(lr){3-7}
&
& Compressed & CODiff$^{*}$ & DiQP$^{*}$ & SIR-Diff$^{*}$ & \textbf{Ours} \\
\midrule
\multirow{2}{*}{AVC}
& 27 & 23.90 & 26.68 & \second{27.06} & 20.89 & \best{27.42} \\
& 42 & 20.51 & \best{21.52} & 20.27 & 20.40 & \second{21.34} \\
\midrule
\multirow{2}{*}{HEVC}
& 27 & 26.03 & \second{25.96} & 25.26 & 21.81 & \best{29.46} \\
& 42 & 22.59 & 23.36 & \second{23.45} & 23.09 & \best{24.34} \\
\midrule
\multirow{2}{*}{VP9}
& 27 & 24.58 & 25.03 & \best{25.86} & 22.13 & \second{25.18} \\
& 42 & 21.56 & 22.85 & \best{26.32} & 22.40 & \second{25.13} \\
\midrule
\multirow{2}{*}{AV1}
& 27 & 23.88 & 25.48 & \second{25.95} & 20.62 & \best{29.32} \\
& 42 & 23.52 & 23.89 & \best{26.25} & 23.06 & \second{25.08} \\
\bottomrule
\end{tabular}%
}
\begin{minipage}{0.92\linewidth}
\footnotesize
\raggedright
$^{*}$ denotes that the model has been finetuned on our training datasets.
\end{minipage}
\end{table}

\begin{table}[!t]
\centering
\caption{Quantitative comparison on the Free dataset under a matched fine-tuning protocol. We report 2D restoration metrics and MEt3R for evaluating image quality and multi-view geometric consistency. The \colorbox{best}{Best} and \colorbox{second}{Second-best} results are highlighted in red and orange, respectively.}
\label{tab:matched_finetune}
\resizebox{\columnwidth}{!}{%
\begin{tabular}{l|ccccccc}
\toprule
         & PSNR$\uparrow$ & SSIM$\uparrow$ & LPIPS$\downarrow$ & NIQE$\downarrow$ & MUSIQ$\uparrow$ & CLIPIQA$\uparrow$ & MEt3R$\downarrow$ \\
\midrule
CODiff$^{*}$   & 23.74 & 0.662 & \second{0.277} & \second{3.40} & \best{69.18} & \best{0.681} & \second{0.135}       \\
DiQP$^{*}$     & \best{27.23} & \second{0.803} & \best{0.225} & 4.06 & 64.99 & 0.393 &  0.136      \\
SIR-Diff$^{*}$ & 20.91 & 0.794 & 0.426 & 3.83 & 60.99 & 0.565 &  0.141 \\
Ours     & \second{23.89} & \best{0.897} & 0.281 & \best{3.26} & \second{68.08} &  \second{0.673} &  \best{0.132}     \\
\bottomrule
\end{tabular}%
}
\begin{minipage}{\linewidth}
\footnotesize
\raggedright
$^{*}$ denotes that the model has been finetuned on our training datasets. 
\end{minipage}
\end{table}

\begin{table}[t]
\centering
\caption{Generalization to the unseen VVC/H.266 codec on the Free dataset. VVC is not included during training. We report PSNR, NIQE, and CLIPIQA, where higher PSNR/CLIPIQA and lower NIQE indicate better restoration quality. The \colorbox{best}{Best} and \colorbox{second}{Second-best} results are highlighted in red and orange, respectively.}
\label{tab:unseen_vvc}
\setlength{\tabcolsep}{5pt}

\resizebox{0.50\columnwidth}{!}{%
\begin{tabular}{lccc}
\toprule
Method
& PSNR $\uparrow$
& NIQE $\downarrow$
& CLIPIQA $\uparrow$ \\
\midrule
CODiff$^{*}$
& \second{23.4970}
& \second{3.3412}
& \second{0.6917} \\
Ours
& \best{23.6228}
& \best{3.2610}
& \best{0.6981} \\
\bottomrule
\end{tabular}%
}
\end{table}

\begin{table*}[h!]
\centering
\caption{Quantitative comparison of 2D restoration methods on the Free dataset. We evaluate various 2D diffusion-based restoration models under different codec and QP settings. The \colorbox{best}{Best} and \colorbox{second}{Second-best} results are highlighted in red and orange, respectively.}
\label{tab:2d_quality_assessment}
\resizebox{\linewidth}{!}{%
\begin{tabular}{ll*{26}{c}}
\toprule
\multirow{2}{*}{Codec} & \multirow{2}{*}{QP}
& \multicolumn{4}{c}{PSNR $\uparrow$}
& \multicolumn{4}{c}{SSIM $\uparrow$}
& \multicolumn{4}{c}{LPIPS $\downarrow$} 
& \multicolumn{4}{c}{NIQE $\downarrow$}
& \multicolumn{4}{c}{MUSIQ $\uparrow$}
& \multicolumn{4}{c}{CLIPIQA $\uparrow$}\\
\cmidrule(lr){3-6} \cmidrule(lr){7-10} \cmidrule(lr){11-14}\cmidrule(lr){15-18} \cmidrule(lr){19-22} \cmidrule(lr){23-26}
& & CODiff & DiQP & SIR-Diff* & \textbf{Ours}
  & CODiff & DiQP & SIR-Diff* & \textbf{Ours}
  & CODiff & DiQP & SIR-Diff* & \textbf{Ours}
  & CODiff & DiQP & SIR-Diff* & \textbf{Ours}
  & CODiff & DiQP & SIR-Diff* & \textbf{Ours}
  & CODiff & DiQP & SIR-Diff* & \textbf{Ours}\\
\midrule
\multirow{2}{*}{AVC}
& 27
& \second{23.29}  & 22.38  & 20.92  & \best{23.45}
& \second{0.896}  & 0.847 & 0.799 & \best{0.906}
& \second{0.356}  & 0.374 & 0.390  & \best{0.224}
& 5.06 & \second{3.37}  & 3.59  & \best{3.10}
& \second{65.60}  & 50.65  & 62.60 & \best{69.84}
& \second{0.647}  & 0.395  & 0.577 & \best{0.704} \\
& 42
& \second{23.29} & 21.65 & 20.73 & \best{23.03}
& \best{0.869}  & 0.802  & 0.778  & \second{0.857}
& \second{0.458}  & 0.484  & 0.461 & \best{0.391}
& 5.37  & 4.47  & \second{4.07}  & \best{3.25}
& 58.48  & 40.66  & \second{61.20}  & \best{65.479}
& 0.529  & 0.315  & \second{0.552}  & \best{0.610} \\
\midrule

\multirow{2}{*}{HEVC}
& 27
& \second{24.42}  & 23.12  & 21.27  & \best{25.98}
& \second{0.897}  & 0.831 & 0.796  & \best{0.916}
& \second{0.355}  & 0.382  & 0.447  & \best{0.218}
& 5.06  & \second{3.49}  & 3.92  & \best{3.24}
& \second{65.02}  & 46.49  & 58.36  & \best{68.39}
& \second{0.634}  & 0.360  & 0.555  & \best{0.689} \\
& 42
& \second{24.50}  & 22.70  & 21.20  & \best{25.61}
& \second{0.889}  & 0.811  & 0.789 & \best{0.900}
& \second{0.409}  & 0.451  & 0.477  & \best{0.320}
& 5.27  & \second{4.06}  & 4.21  & \best{3.42}
& \second{62.54}  & 41.93  & 58.18  & \best{67.11}
& \second{0.581}  & 0.330  & 0.547  & \best{0.640} \\
\midrule

\multirow{2}{*}{AV1}
& 27
& \second{23.30}  & 22.34  & 20.91  & \best{23.74}
& \second{0.897}  & 0.845  & 0.798  & \best{0.906}
& \second{0.359} & 0.382 & 0.397  & \best{0.236}
& 5.05  & \second{3.47}  & 3.62 & \best{3.20}
& \second{65.82}  & 50.48  & 62.89 & \best{69.51}
& \second{0.645}  & 0.401  & 0.577  & \best{0.709} \\
& 42
& \second{23.45}  & 21.99  & 20.83  & \best{23.75}
& \best{0.894}  & 0.827  & 0.793  & \second{0.893}
& \second{0.408} & 0.444  & 0.438 & \best{0.327}
& 5.22  & \second{4.08}  & 3.90  & \best{3.35}
& \second{63.81}  & 45.07  & 62.18  & \best{67.13}
& \second{0.597}  & 0.362   & 0.564  & \best{0.658} \\
\midrule

\multirow{2}{*}{VP9}
& 27
& \second{23.26}  & 22.39  & 20.91  & \best{23.40}
& \second{0.896} & 0.847  & 0.799  & \best{0.906}
& \second{0.357}  & 0.372  & 0.388  & \best{0.219}
& 5.03  & \second{3.40}  & 3.57  & \best{3.20}
& \second{65.65}  & 50.97  & 62.82  & \best{69.95}
& \second{0.647} & 0.407  & 0.578  & \best{0.713} \\
& 42
& \second{23.41}  & 22.06  & 20.85  & \best{23.72}
& \best{0.893}  & \second{0.830} & 0.793  & \best{0.893}
& \second{0.402}  & 0.433 & 0.431  & \best{0.314}
& 5.18  & 3.92 & \second{3.85} & \best{3.24}
& \second{63.81}  & 45.88  & 62.08 & \best{67.53}
& \second{0.598}  & 0.369  & 0.563  & \best{0.662} \\
\bottomrule
\end{tabular}
}
\begin{minipage}{\linewidth}
\footnotesize
\raggedright
$^{*}$ denotes that the model has been finetuned on our training datasets. 
\end{minipage}
\end{table*}

\subsection{2D evaluation}
\begin{figure*}[p]
  \centering
  \includegraphics[width=\textwidth]{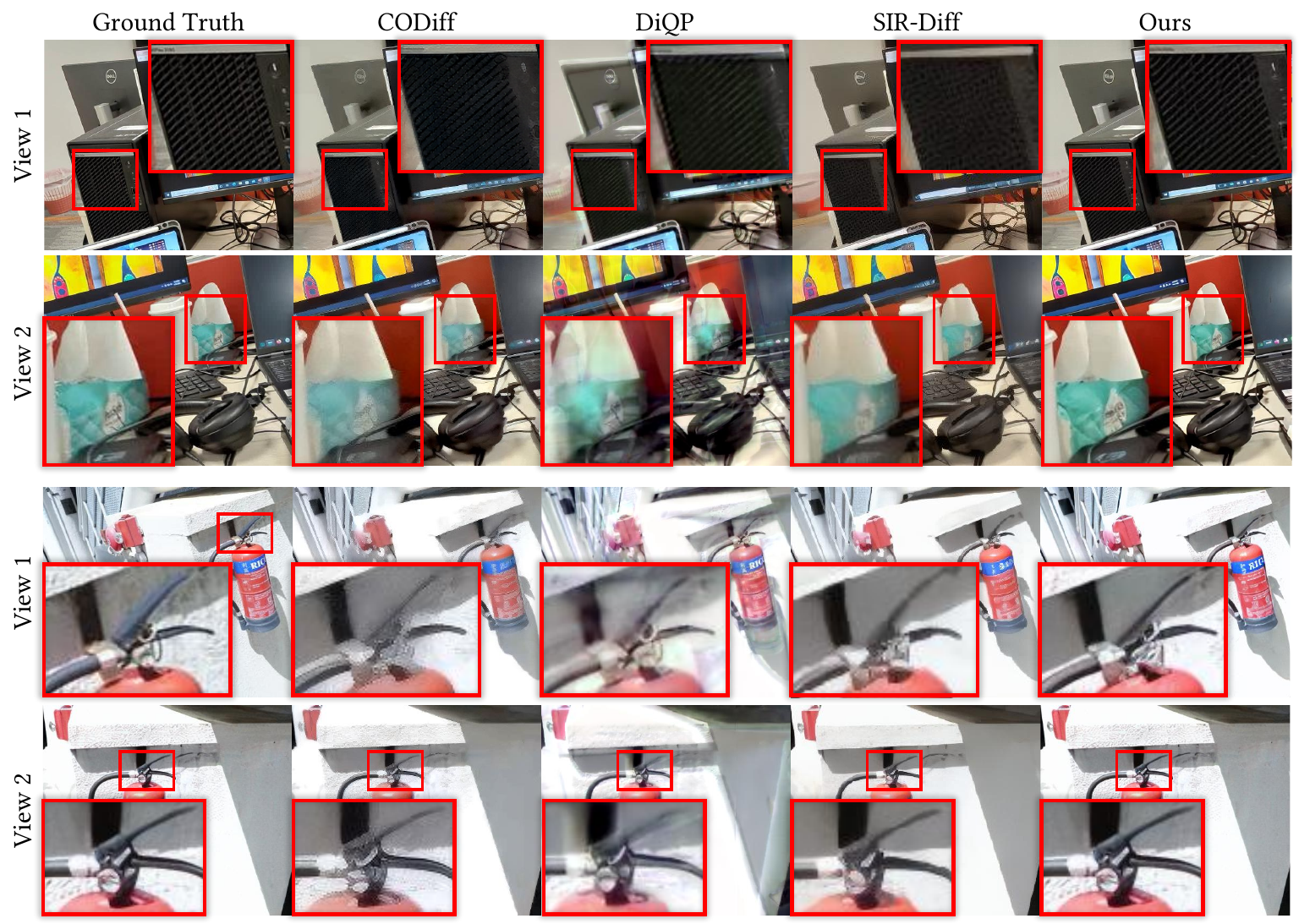}
  \caption{Qualitative comparison on the Free dataset. The top rows correspond to an indoor scene (Lab), and the bottom rows correspond to an outdoor scene (Sky). The Lab scene is compressed using HEVC with QP 42, while the Sky scene is compressed using AV1 with QP 42. Highlighted regions reveal that our method better preserves fine details and structures under severe compression.}
  \label{fig:2D_visual_figure}
\end{figure*}

\paragraph{Quality Assessments.}
We compare CompMVR with CODiff~\cite{guo2025compression}, DiQP~\cite{dehaghi2025reversing}, and the multi-view restoration model SIR-Diff~\cite{mao2025sir}. Tab.~\ref{tab:2d_quality_assessment} shows that our method consistently achieves the best or second-best performance across 2D quality metrics under various codec and QP settings and remains robust under severe compression (e.g., QP 42). Fig.~\ref{fig:2D_visual_figure} further shows that existing methods often leave compression artifacts or over-smooth high-frequency structures, whereas CompMVR better preserves structural details and visual fidelity. In the sky scene, competing methods restore the same object inconsistently across views through artifact amplification or excessive smoothing, while our method produces more consistent restorations. Together, these results show that compression-aware conditioning and geometry-aware consistency modeling enable high-quality, view-consistent restoration. On unseen VVC/H.266~\cite{bross2021overview}, CompMVR also improves PSNR, NIQE, and CLIPIQA over CODiff (Tab.~\ref{tab:unseen_vvc}), demonstrating generalization beyond the training codecs.

\subsection{Ablation Studies}
\begin{figure}
  \centering
  \includegraphics[width=0.90\columnwidth]{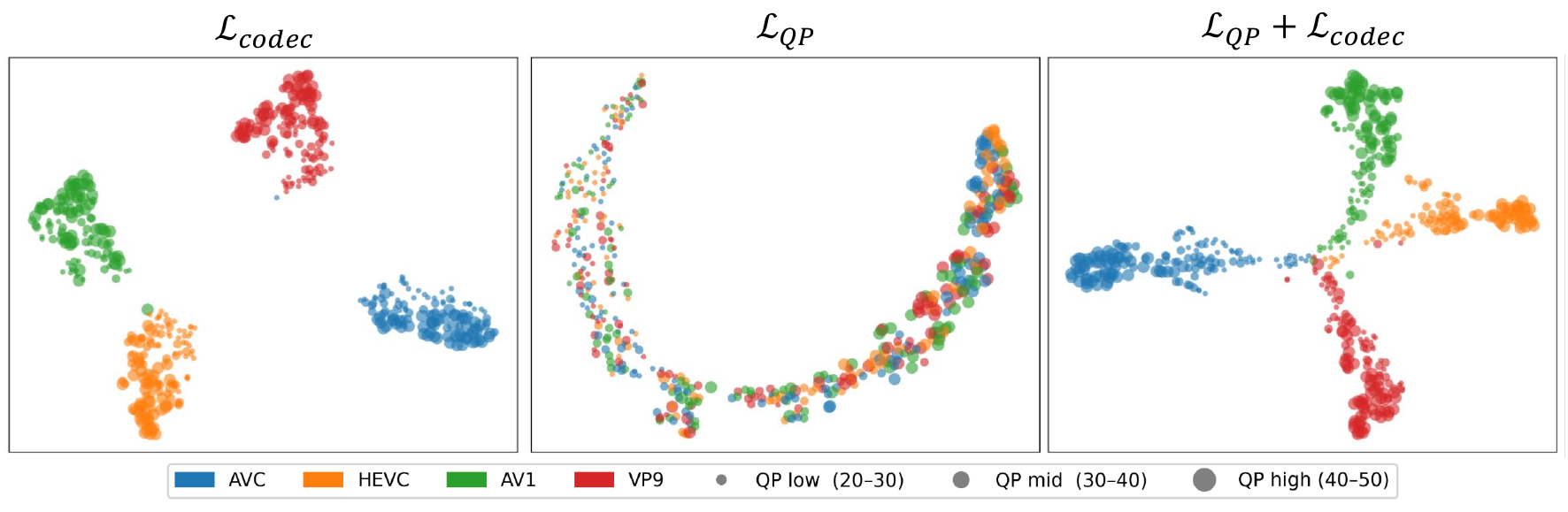}
  \caption{t-SNE visualization of the Stage 1 latent representations learned by using $\mathcal{L}_{\text{codec}}$ and $\mathcal{L}_{\text{QP}}$, and their combination, respectively. Color denotes codec type (AVC / HEVC / AV1 / VP9) and point size denotes QP level.}
  \label{fig:tsne_cpe}
\end{figure}

\paragraph{Ablation of the Compression-Aware Prior}
As shown in Tab.~\ref{tab:abl1}, $\mathcal{L}_{\text{QP}}$ provides stronger guidance than $\mathcal{L}_{\text{codec}}$ alone, while their combination achieves the best or second-best results across all metrics. The t-SNE visualization in Fig.~\ref{fig:tsne_cpe} shows that $\mathcal{L}_{\text{codec}}$ separates codec types, whereas $\mathcal{L}_{\text{QP}}$ organizes representations by compression severity, with both capturing complementary codec- and QP-dependent structures. Replacing the learned Stage-1 prior with a null embedding increases Sampson error from 0.078 to 0.081 and MEt3R from 0.1357 to 0.1361 (Tab.~\ref{tab:prior_geometry}), while random conditioning shows consistent degradation across multiple geometric metrics (Tab.~\ref{tab:random_prior}). These results indicate that the learned compression prior provides a consistent, though moderate, contribution to geometric reliability.

\begin{table}[t]
\centering
\caption{Ablation study on Stage 1 supervision losses, $\mathcal{L}_{\text{codec}}$ and $\mathcal{L}_{\text{QP}}$, evaluated by the final Stage 2 restoration performance. The \colorbox{best}{Best} and \colorbox{second}{Second-best} results are highlighted in red and orange, respectively.}
\label{tab:abl1}
\resizebox{0.75\columnwidth}{!}{%
\begin{tabular}{cc|cccccc}
\toprule
$\mathcal{L}_{\text{codec}}$ & $\mathcal{L}_{\text{QP}}$
& PSNR$\uparrow$ & SSIM$\uparrow$ & LPIPS$\downarrow$
& NIQE$\downarrow$ & MUSIQ$\uparrow$ & CLIPIQA$\uparrow$ \\
\midrule
\checkmark &           & 23.49 & 0.887 & 0.304 & 3.40 & 63.69 & 0.613 \\
           & \checkmark & \second{23.77} & \second{0.896} & \second{0.282} & \second{3.32} & \best{68.42} & \best{0.676} \\
\checkmark & \checkmark & \best{23.89} & \best{0.897} & \best{0.281} & \best{3.26} & \second{68.08} & \second{0.673} \\
\bottomrule
\end{tabular}%
}
\end{table}

\begin{table}[t]
\centering
\caption{Geometric ablation of the learned Stage-1 compression prior. The \colorbox{best}{Best} and \colorbox{second}{Second-best} results are highlighted in red and orange, respectively.}
\label{tab:prior_geometry}
\small
\setlength{\tabcolsep}{5pt}
\resizebox{0.50\columnwidth}{!}{%
\begin{tabular}{lcc}
\toprule
Conditioning
& Sampson $\downarrow$
& MEt3R $\downarrow$ \\
\midrule
Null embedding
& \second{0.081} & \second{0.1361} \\
Learned prior
& \best{0.078} & \best{0.1357} \\
\bottomrule
\end{tabular}%
}
\end{table}

\begin{table}[t]
\centering
\caption{Robustness analysis of the learned compression prior against random conditioning. The \colorbox{best}{Best} and \colorbox{second}{Second-best} results are highlighted in red and orange, respectively.}
\label{tab:random_prior}
\resizebox{\columnwidth}{!}{%
\begin{tabular}{lccccc}
\toprule
Conditioning
& Inlier Ratio $\uparrow$
& RANSAC Inliers $\uparrow$
& Sampson $\downarrow$
& AUC@5 $\uparrow$
& MEt3R $\downarrow$ \\
\midrule
Random
& \second{0.7747} & \second{97.7} & \second{0.0960} & \second{0.5164} & \second{0.1494} \\
Learned
& \best{0.8047} & \best{130.1} & \best{0.0806} & \best{0.5275} & \best{0.1404} \\
\bottomrule
\end{tabular}}
\end{table}

\aptLtoX{\begin{table}[t]
\caption{Ablation study on Stage 2 components, including the compression prior extracted from Stage 1, the compression artifact estimator (CAE), and the cross-view correspondence loss $\mathcal{L}_{\text{corr}}$. The \colorbox{best}{Best} and \colorbox{second}{Second-best} results are highlighted in red and orange, respectively.}
\begin{tabular}{ccc|cccccc}
\hline  
Compression Prior & CAE & $\mathcal{L}_{\text{corr}}$ 
& PSNR$\uparrow$ & SSIM$\uparrow$ & LPIPS$\downarrow$ 
& NIQE$\downarrow$ & MUSIQ$\uparrow$ & CLIPIQA$\uparrow$ \\
\hline
\checkmark &           &           & 23.79 & 0.893 & 0.305 & \best{3.16} & 62.16 & 0.595 \\
\checkmark & \checkmark &           & 23.71 & 0.892 & 0.293 & 3.27 & 67.42 & \second{0.672} \\
\checkmark &           & \checkmark & 23.82 & 0.896 & 0.287 & 3.39 & \second{67.68} & 0.665 \\
 & \checkmark & \checkmark & \second{23.87} & \best{0.898} & \best{0.267} & 3.54 & 67.11 & 0.669 \\
\checkmark & \checkmark & \checkmark & \best{23.89} & \second{0.897} & \second{0.281} & \second{3.26} & \best{68.08} & \best{0.673} \\
\hline
\end{tabular}
\end{table}
\begin{table}[h!]
\caption{Geometric ablation of the cross-view correspondence loss $\mathcal{L}_{\mathrm{corr}}$. The \colorbox{best}{Best} and \colorbox{second}{Second-best} results are highlighted in red and orange, respectively.}
\begin{tabular}{lcccc}
\hline
Variant
& Inlier Ratio $\uparrow$
& Sampson $\downarrow$
& AUC@5 $\uparrow$
& MEt3R $\downarrow$ \\
\hline
w/o $\mathcal{L}_{\mathrm{corr}}$
& \second{0.801} & \second{0.082} & \second{0.516} & \second{0.138} \\
Full
& \best{0.805} & \best{0.078} & \best{0.531} & \best{0.132} \\
\hline
\end{tabular}
\end{table}}{\begin{table}[t]
\centering
\caption{Ablation study on Stage 2 components, including the compression prior extracted from Stage 1, the compression artifact estimator (CAE), and the cross-view correspondence loss $\mathcal{L}_{\text{corr}}$. The \colorbox{best}{Best} and \colorbox{second}{Second-best} results are highlighted in red and orange, respectively.}
\label{tab:abl2}
\resizebox{\columnwidth}{!}{%
\begin{tabular}{ccc|cccccc}
\toprule  %
Compression Prior & CAE & $\mathcal{L}_{\text{corr}}$ 
& PSNR$\uparrow$ & SSIM$\uparrow$ & LPIPS$\downarrow$ 
& NIQE$\downarrow$ & MUSIQ$\uparrow$ & CLIPIQA$\uparrow$ \\
\midrule
\checkmark &           &           & 23.79 & 0.893 & 0.305 & \best{3.16} & 62.16 & 0.595 \\
\checkmark & \checkmark &           & 23.71 & 0.892 & 0.293 & 3.27 & 67.42 & \second{0.672} \\
\checkmark &           & \checkmark & 23.82 & 0.896 & 0.287 & 3.39 & \second{67.68} & 0.665 \\
 & \checkmark & \checkmark & \second{23.87} & \best{0.898} & \best{0.267} & 3.54 & 67.11 & 0.669 \\
\checkmark & \checkmark & \checkmark & \best{23.89} & \second{0.897} & \second{0.281} & \second{3.26} & \best{68.08} & \best{0.673} \\
\bottomrule
\end{tabular}%
}
%\end{table}
%\begin{table}[h!]
\centering
\caption{Geometric ablation of the cross-view correspondence loss $\mathcal{L}_{\mathrm{corr}}$. The \colorbox{best}{Best} and \colorbox{second}{Second-best} results are highlighted in red and orange, respectively.}
\label{tab:lcorr_geometry}
\resizebox{0.80\columnwidth}{!}{%
\begin{tabular}{lcccc}
\toprule
Variant
& Inlier Ratio $\uparrow$
& Sampson $\downarrow$
& AUC@5 $\uparrow$
& MEt3R $\downarrow$ \\
\midrule
w/o $\mathcal{L}_{\mathrm{corr}}$
& \second{0.801} & \second{0.082} & \second{0.516} & \second{0.138} \\
Full
& \best{0.805} & \best{0.078} & \best{0.531} & \best{0.132} \\
\bottomrule
\end{tabular}}
\end{table}
}

\paragraph{Ablation on Stage 2 Components}
We evaluate the impact of each Stage 2 component, including the compression prior from Stage 1, the compression artifact estimator (CAE) and the cross-view correspondence loss $\mathcal{L}_{\text{corr}}$. As shown in Tab.~\ref{tab:abl2}, the compression prior alone provides competitive performance, while combining it with either CAE or $\mathcal{L}_{\text{corr}}$ generally improves perceptual quality. However, the gains from the compression prior are modest and metric-dependent, indicating that it provides a complementary rather than dominant contribution to Stage 2. Nevertheless, using all three components yields the most balanced performance, achieving the best or second-best results across most metrics. The geometry-focused ablation in Tab.~\ref{tab:lcorr_geometry} further isolates the effect of correspondence supervision. Adding $\mathcal{L}_{\mathrm{corr}}$ improves the inlier ratio from 0.801 to 0.805, reduces the Sampson error from 0.082 to 0.078, improves AUC@5 from 0.516 to 0.531, and reduces MEt3R from 0.138 to 0.132. These results confirm the geometry-side contribution of correspondence supervision beyond image-level restoration.

\begin{table}[t]
\centering
\caption{Computational complexity of the compared restoration methods. We report the number of trainable parameters, forward inference time, and peak GPU memory usage.}
\label{tab:complexity}
\setlength{\tabcolsep}{4pt}
\renewcommand{\arraystretch}{0.95}
\resizebox{0.70\columnwidth}{!}{%
\begin{tabular}{lccc}
\toprule
Method
& Params (M) $\downarrow$
& Time (s) $\downarrow$
& Memory (GB) $\downarrow$ \\
\midrule
DiQP & 79.373 & 0.422 & 1.02 \\
CODiff & 93.113 & 0.699 & 2.89 \\
SIR-Diff & 815.545 & 0.826 & 2.48 \\
Ours & 276.171 & 1.073 & 11.01 \\
\bottomrule
\end{tabular}
}
\end{table}

\section{Discussions}
\paragraph{Limitations}
As reported in Tab.~\ref{tab:complexity}, our framework requires additional computation and GPU memory, and it is not yet a fully feed-forward pipeline. Developing a faster and more memory-efficient feed-forward version could be a promising future direction. In addition, although we evaluate on several representative downstream tasks, direct applicability to untested 3D tasks requires further validation. Finally, our results demonstrate the importance of coding priors, but also reveal a limitation: different codecs exhibit different artifacts and coding structures. Thus, codec-adaptive priors, especially for codecs such as AV1~\cite{chen2020overview} whose coding tools differ from HEVC~\cite{sullivan2012overview}, are needed for stronger codec generalization. Additional failure cases and region-wise analyses over textured, edge, and smooth regions are provided in the supplementary material to examine restoration fidelity and plausible but inaccurate details.

\paragraph{Conclusion}
In this paper, we introduce \textbf{CompMVR}, a compression-aware multi-view diffusion restoration framework designed to improve geometric reliability for downstream 3D reconstruction. Our method leverages coding priors and cross-view supervision to restore geometrically reliable multi-view inputs. Through extensive experiments across multiple codecs and QP settings, our results demonstrate that CompMVR improves novel-view synthesis and multiple downstream geometric tasks. Our results show that compression artifacts affect downstream 3D reconstruction, and that the restoration process should consider geometric reliability when targeting 3D vision applications.

\begin{acks} 
This research was supported in part by the National Research Foundation of Korea (NRF) grant funded by the Korean government (MSIT) (RS-2025-00520308, 50\%), in part by the Regional Innovation System \& Education (RISE) Glocal 30 program through the Daegu RISE Center, funded by the Ministry of Education (MOE) and Daegu, Republic of Korea (2025-RISE-03-001, 25\%), and in part by Culture, Sports and Tourism R\&D Program through the Korea Creative Content Agency, funded by the Ministry of Culture, Sports and Tourism in 2026 (Project Name: Nurturing Global Technical Experts for Performance Evaluation of Multimodal Content Copyright Core Technologies, Project Number: RS-2026-2552393, Contribution Rate: 25\%).
\end{acks}

\clearpage

\clearpage

\clearpage

\bibliographystyle{ACM-Reference-Format}
\bibliography{reference}
\end{document}